\pdfoutput=1

\documentclass[11pt]{article}

\usepackage[final]{acl}
\usepackage{xspace}

\usepackage{times}
\usepackage{latexsym}

\usepackage{smile}
\usepackage{amsfonts}
\usepackage{amssymb}
\usepackage[T1]{fontenc}

\usepackage[utf8]{inputenc}

\usepackage{microtype}

\usepackage{xcolor, colortbl}  
\usepackage{inconsolata}
\usepackage{multirow}
\usepackage{subfigure}
\usepackage{algorithm}
\usepackage{algorithmic}
\usepackage{listings}
\usepackage{pythonhighlight}
\usepackage{fancyvrb}
\usepackage{enumitem}

\usepackage{graphicx}
\usepackage{array}

\usepackage{subfigure}
\usepackage{caption}
\usepackage{listing}

\lstdefinestyle{mystyle}{
  basicstyle=\ttfamily,
  frame=single,
  breaklines=true,
  breakindent=0pt,
  backgroundcolor=\color{gray!10}, 
}

\definecolor{green}{HTML}{009B55}
\definecolor{lightcoral}{HTML}{F08080}

\title{\ours{}: Self-Improving Retrieval-Augmented Generation for \\ Adapting Large Language Models to Specialized Domains}

\newcommand{\ours}{{SimRAG}\xspace}

\author{Ran Xu$^{1,2}\thanks{Work done during an internship at Amazon.}$, Hui Liu$^2$, Sreyashi Nag$^2$, Zhenwei Dai$^2$, Yaochen Xie$^2$, Xianfeng Tang$^2$,\\ \bf Chen Luo$^2$, Yang Li$^2$, Joyce C. Ho$^1$, Carl Yang$^1$, Qi He$^2$ \\
\vspace{2pt}
$^1$ Emory University~~~~$^2$ Amazon \\
\texttt{\{ran.xu,joyce.c.ho,j.carlyang\}@emory.edu}, \texttt{liunhu@amazon.com}
}

\begin{document}
\maketitle
\begin{abstract}
Retrieval-augmented generation (RAG) enhances the question answering (QA) abilities of large language models (LLMs) by integrating external knowledge. However, adapting general-purpose RAG systems to specialized fields such as science and medicine poses unique challenges due to distribution shifts and limited access to domain-specific data. To tackle this, we propose \ours{}, a self-training approach that equips the LLM with joint capabilities of question answering and question generation for domain adaptation. Our method first fine-tunes the LLM on instruction-following, question-answering, and search-related data. Then, it prompts the same LLM to generate diverse domain-relevant questions from unlabeled corpora, with an additional filtering strategy to retain high-quality synthetic examples. 
By leveraging these self-generated synthetic examples, the LLM can improve their performance on domain-specific RAG tasks. 
Experiments on 11 datasets, spanning two backbone sizes and three domains, demonstrate that \ours{} outperforms baselines by 1.2\%--8.6\%.

\end{abstract}

\section{Introduction}

Retrieval-augmented generation (RAG)~\citep{lewis2020retrieval,gao2023retrieval,gutierrez2024hipporag,asai2024selfrag} is a powerful technique that enhances large language models (LLMs) for various knowledge-intensive tasks such as question answering (QA) by incorporating external knowledge sources. 
This method not only customizes responses to handle long-tail knowledge but also avoids the need for costly model retraining~\citep{ovadia2023fine}. Additionally, RAG helps reduce the issue of LLM hallucination by ensuring responses are grounded in relevant evidence~\citep{shuster-etal-2021-retrieval-augmentation}, thereby improving the overall accuracy and reliability of LLM outputs.

While extensive research has focused on developing effective~\citep{asai2024selfrag,lin2024radit,liu2024chatqa} and efficient~\citep{xu2024recomp} RAG systems for general-domain QA tasks, adapting RAG to specialized domains for LLMs poses significant challenges. 
These models often struggle with distribution shifts and fail to accurately extract information from domain-specific contexts~\citep{miller2020effect,liu-etal-2022-challenges}. 
Moreover, directly using black-box LLMs~\citep{gpt4,claude2,wang2023augmenting} in specialized domains raises concerns about privacy when dealing with sensitive proprietary data. 
It is essential to fine-tune LLMs on domain-relevant QA tasks to unlock the full potential of LLM-based RAG systems in specialized domains.

Despite the critical need for domain-specific fine-tuning, 
the primary challenge lies in the acquisition of high-quality
fine-tuning data towards RAG applications. 
Prior works rely on continuous pretraining~\citep{chen2023meditron,zhang2024sciglm} on specialized corpora or fine-tuning on domain-specific instruction-tuning data~\citep{pmcllama,wadden2024sciriff}. However, the mismatch between these general-purpose tasks and domain-specific QA hinders their effectiveness. 
More recently, several approaches~\citep{liu2024chatqa,synqa,zhang2024raft} use synthetic data from powerful LLMs (e.g., GPT-4) to create QA fine-tuning datasets. While promising, these methods are costly, inefficient, and lack explicit quality control over the generated outputs. Additionally, the direct use of proprietary corpora with black-box LLMs introduces privacy concerns, making these methods unsuitable for sensitive domains.

To tackle the data scarcity issue mentioned above, we propose \ours{}\footnote{\underline{S}elf-\underline{im}proving \underline{R}etrieval-\underline{A}ugmented \underline{G}eneration.}, a self-improving approach to harness the LLMs' own capabilities to generate pseudo-labeled data for domain adaptative question answering. 
Our method is inspired by the success of self-training in LLM development, where models are refined using synthetic examples generated from unlabeled corpora \citep{wang-etal-2022-gpl,li2024selfalignment}. However, for RAG applications, special considerations are needed to adapt LLMs for generating questions that require external context to answer. The core objective of \ours{} is to fine-tune a single LLM to perform two complementary tasks: \emph{question answering with context} and \emph{question generation from context}. Both tasks involve extracting and summarizing relevant information from the context, allowing them to mutually reinforce each other.

Specifically, we design a two-stage procedure to adapt LLMs for domain QA, we first fine-tune LLMs on \emph{instruction-following}, \emph{question answering}, and \emph{question generation} data from general-domains. This step equips LLMs with basic instruction-following and context utilization skills. 
Then, to specialize the model for domain-specific tasks, we then harness unlabeled domain corpora, prompting the same LLM to generate high-quality QA pairs grounded in the context of these specialized domains.
To further \emph{enhance the quality} of synthetic pairs, we incorporate multiple task types to improve the model's generalization capabilities, combined with round-trip consistency filtering technique~\citep{bartolo-etal-2021-improving} to preserve generated QA pairs only when the original context is retrieved among top results.  
With these pseudo-labeled (\emph{question}, \emph{passage}, \emph{answer}) tuples generated by LLMs, we continuously fine-tune the models with those synthetic examples. 
This pipeline allows the LLM to progressively refine its output on synthetic pairs, thus adapting itself towards domain-specific QA applications.

We conduct experiments on three different domains spanning from biomedical, natural/social sciences, and computer science (CS), where we observe \ours{} consistently achieve better performance than other domain-specific LLMs and general-domain retrieval-augmented LMs. Qualitative studies highlight the benefits of joint training in question answering and generation, along with diverse, denoised QA pairs.

Our contribution can be summarized as follows:
\begin{itemize}[leftmargin=0.5cm]
    \item We propose \ours{}, a RAG framework that enhances LLM's capability for question answering on specialized domains.  
    \item We design a novel instruction fine-tuning approach that enables LLMs to perform both question answering and question generation. This joint capability facilitates self-improvement through self-training on generated synthetic data, leading to enhanced model performance.
    \item We validate our approach with empirical studies across 11 datasets from three distinct domains, demonstrating that \ours{} outperforms baseline models by 1.2\%--8.6\%.
\end{itemize}
\section{Related Work}
\vspace{-.5ex}
\label{sec:related_work}
\noindent \textbf{Retrieval-augmented generation.} RAG has emerged as a powerful tool in knowledge-intensive NLP tasks such as language modeling~\citep{borgeaud2022improving} and question answering~\citep{lewis2020retrieval,shi2023replug}. The typical approach involves integrating a retriever with the LLM generator and designing a fine-tuning process to align the retriever with LLM capabilities. 
To further refine RAG, recent research explored various enhancements. These include developing dynamical retrieval processes to refine the relevance of fetched content~\citep{jiang-etal-2023-active, jeong-etal-2024-adaptive,su-etal-2024-dragin}, and filtering out irrelevant contexts to robustify RAG~\citep{robustlm,yu2024rankrag,chain_of_note,wang2024blendfilter}. Additionally, several studies have developed instruction-tuning methods aimed specifically at improving search and RAG capabilities of LLMs~\citep{liu2024chatqa,lin2024radit,dong2024understand,wei2024instructrag}.

\noindent \textbf{Self-training.} 
Self-training (or Pseudo-Labeling) is one of the earliest approaches to semi-supervised learning~\citep{rosenberg2005semi}.
The method uses a teacher model to generate new labels on which a student model is fitted. 
Self-training has been widely adopted for various NLP tasks including text classification~\citep{du2021self}, natural language understanding~\citep{vu2021strata} and  ranking~\citep{wang-etal-2022-gpl}. 
Recently, the idea of self-training has also been applied to LLM instruction fine-tuning~\citep{yuan2024self,li2024selfalignment},  reasoning~\citep{pang2024iterative}, and alignment~\citep{gulcehre2023reinforced}, yet to the best of our knowledge, this pipeline has not been widely explored for RAG applications. 
The major drawback of self-training is that it is vulnerable to label noise~\cite{arazo2020pseudo}. 
There are several approaches to stabilize the self-training, with sample selection~\citep{li2024selfalignment} and reweighting~\citep{wang2021meta} strategies.

\begin{figure*}
    \centering    \includegraphics[width=0.98\linewidth]{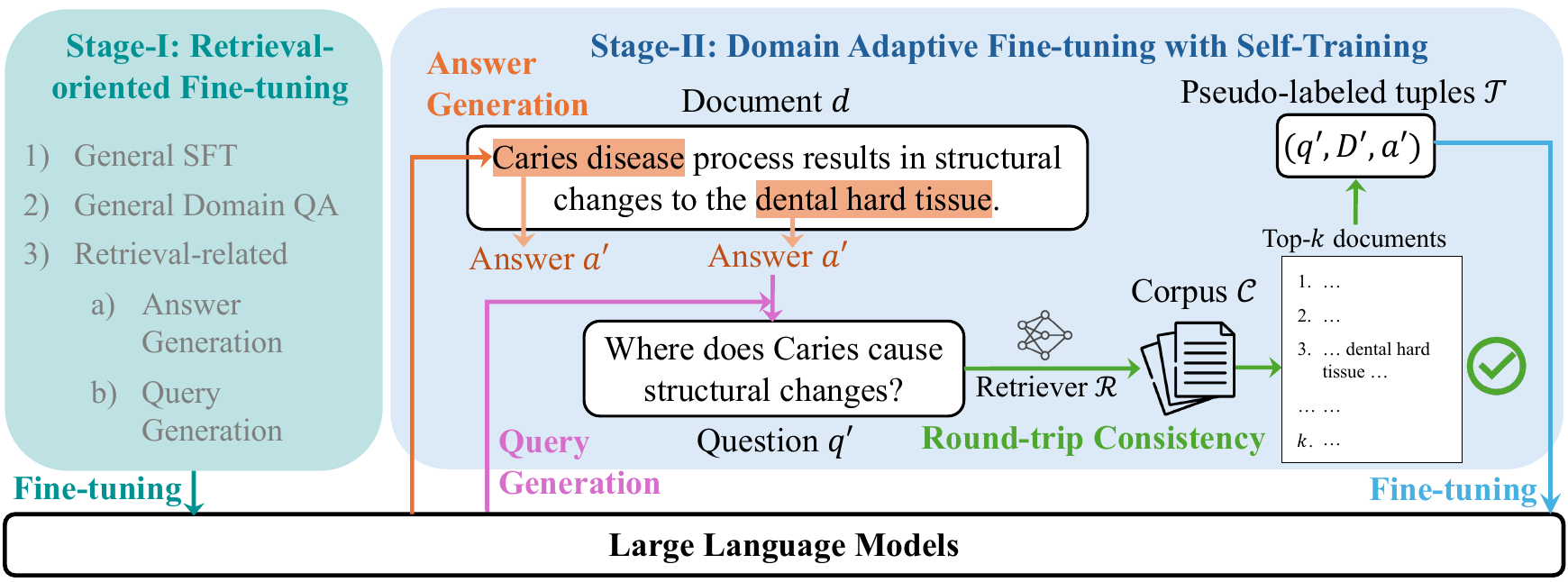}
    \caption{Two-stage fine-tuning framework for our proposed method \ours{}. The model is first fine-tuned on retrieval-related data. Then, it generates pseudo-labeled tuples by first extracting candidate answers from the corpus, and then generating candidate questions conditioned on both document and answer. The LLM is further fine-tuned on pseudo-labeled examples filtered with round-trip consistency. }
    \label{fig:framework}
\end{figure*}

\noindent \textbf{Domain-specific LLMs.} Most domain-specific LLMs rely on continuous pretraining~\citep{labrak2024biomistral,chen2023meditron,xu2024bmretriever} or domain-specific fine-tuning~\citep{pmcllama,zhang2024sciglm,zhang2023alpacare,wadden2024sciriff,shi2024medadapter}, with little focus on adapting models for domain-specific RAG settings. Relevant works~\citep{zhang2024raft,synqa} use strong GPT models for synthetic data generation in RAG scenarios. In contrast, \ours{} leverages the same LLM for both question generation and answering, enabling self-improvement and offering a more cost-effective approach for adapting LLMs to domain-specific QA tasks.


\section{Methodology}
\subsection{Problem Setup}
In a RAG problem, we aim to generate answers for queries based on a set of supporting documents or contexts. Specifically, for a query $q$, an retriever $\cR$ is utilized to retrieve top-$k$ most relevant contexts $\cD=\{d_1,d_2,...,d_k\}$ from a large corpus $\cC$. The LLM $\cM_{\theta}$ then generates an answer $a$ to the query $q$ based on the retrieved context $\cD$.

In this work, we aim to improve the LLM's QA capability in RAG system towards  specialized domains where only unlabeled corpus $\cC$ is available. As shown in Figure~\ref{fig:framework}, our approach first learns from retrieval-oriented instruction data in the general domain in Stage-I and then augments $\cT$ with pseudo-labeled $\cT'=(q', \cD', a')$ tuples in Stage-II, where $\cD'$ is sampled from the specialized domain $\cC$ for self-training.
The overall objective of our study is to adapt the LLM $\cM_{\theta}$ to specialized domains with $\cT\cup\cT'$.

\subsection{Stage-I: Retrieval-oriented fine-tuning}

To start with, we leverage instruction fine-tuned  LLMs as the backbone (e.g. \url{meta-llama/Meta-Llama-3-8B-Instruct}). 
Although these models have been instruction finetuned, 
they still exhibit a deficiency in leveraging context information to answer domain-specific questions.
To improve the their abilities on knowledge-intensive tasks, we fine-tune the LLM with retrieval-oriented tasks. Specifically, we follow~\citet{lin2024radit,liu2024chatqa} and leverage the training data blend that consists of the following components:

(1) \textbf{General Instruction Fine-tuning (SFT) data.} To help maintain the models' ability to comprehend and follow instructions, we leverage the SFT data including OpenAssistant~\citep{köpf2023openassistant}, Dolly~\citep{DatabricksBlog2023DollyV2}, SODA~\citep{kim2022soda}, ELI5~\citep{fan2019eli5}, Self-Instruct~\citep{wang2023self}, and Unnatural Instructions~\citep{honovich2022unnatural}. Note that we make sure there is no overlap between SFT data and test data from target tasks.

(2) \textbf{General domain Context-aware QA data.} To bolster the LLMs' general RAG skills of generating accurate answers grounded in relevant contexts, we fine-tune them on a diverse array of general domain question-answering datasets. This includes DROP~\citep{dua2019drop}, 
NQ~\citep{nq}, 
Squad~\citep{rajpurkar2016squad},
NarrativeQA~\citep{kovcisky2018narrativeqa}, 
Quoref~\citep{dasigi2019quoref},  ROPES~\citep{lin2019reasoning},
OpenbookQA~\citep{mihaylov2018can}, LogiQA~\citep{liu2020logiqa}, TAT-QA~\citep{zhu2021tat}, WebGLM~\citep{liu2023webglm}, StrategyQA~\citep{geva2021did}, BoolQ~\citep{clark2019boolq}, FaVIQ~\citep{park2021faviq} and FEVER~\citep{thorne2018fever} datasets, 
where for each sample, a query $q$ and its relevant context $\cD$ is given, and the LLM is trained to generate answer $a$ to the query.

(3) \textbf{General Retrieval-related Data:} To better generate high-quality pseudo-labeled QA samples in the next stage, we incorporate retrieval-related data to improve two specific skills of LLMs:
(a) \textit{Answer Generation}: where a grounding document is given, and the LLMs are trained to generate candidate spans from the context that are likely to be answers to some questions. In this part, we incorporate Squad 1.1 and 2.0 versions~\citep{rajpurkar2016squad}, DROP~\citep{dua2019drop} and WebQuestions~\citep{berant2013semantic} datasets.
(b) \textit{Query Generation}: where an answer and its grounding document are given, and the LLMs are trained to generate a query based on the document and answer. In this part, we leverage NQ~\citep{nq}, Squad 1.1~\citep{rajpurkar2016squad}, StrategyQA~\citep{geva2021did}, WebQuestions~\citep{berant2013semantic}, FaVIQ~\citep{park2021faviq} and FEVER~\citep{thorne2018fever} datasets.

The details for each dataset (e.g. the instruct format and the amount of data used) are deferred to Appendix~\ref{apd:data_ratio}. 
For each sample in the fine-tuning dataset, we adopt a standard instruction finetuning objective, computing the loss exclusively on the tokens of the assistant's response.

\subsection{Stage-II: Domain Adaptive Fine-tuning}

The model after Stage-I is only trained in the general domains. When directly adopting them to specialized applications, the performance can still be suboptimal due to the distribution shift issue~\citep{miller2020effect}. 
To tailor the LLMs for specialized domains and address the scarcity of labeled data in these areas, we employ a self-training approach leveraging domain-specific unlabeled corpora.
This method capitalizes on the model's enhanced capabilities from the previous retrieval-augmented fine-tuning stage. We utilize the fine-tuned LLM to generate pseudo-labeled training samples $\cT'=(q', \cD', a')$ by creating queries grounded in the unlabeled text and gathering the corresponding retrieved documents.

Specifically, we conduct a two-step procedure to synthesize additional training data, which corresponds to the two skills learned in Stage-I:
(a) \textit{Answer Generation}: for each document $d_i \in \mathcal{C}$, where $\mathcal{C}$ is the unlabeled corpus, we prompt our fine-tuned LLMs to generate several candidate spans ${a_i^1, a_i^2, \dots, a_i^m}$ that are likely to be answers to some questions. Formally, the model generates $a_i^j \sim p_{\theta}(\cdot | d_i)$ for $j = 1, \dots, m$. 
(b) \textit{Answer-conditioned Query Generation}: for each candidate answer $a_i^j$ and its corresponding document $d_i$, we prompt the fine-tuned LLM again to generate candidate questions $q_i^{j} \sim p_{\theta}(\cdot | a_i^j, d_i)$, with $a_i^j$ as the ground truth answer and $d_i$ as the supporting context. This gives us the pseudo-labeled query-answer pair $(q_i^{j}, a_i^j)$ based on the context $d_i$.

During this process, we adopt two additional strategies, namely \emph{diverse question generation} and \emph{data filtering}, to further improve the quality of the synthetic pairs. For diverse question generation, we prompt the LLM to create various types of questions, including \emph{short-span question-answering}, \emph{multiple-choice question-answering}, and \emph{claim verification} tasks. While short-span questions follow the same pipeline as previously described, multiple-choice questions are constructed by using alternative candidate answers from the same unlabeled corpus in step (a) as incorrect options. Claim verification, on the other hand, bypasses the answer generation step; instead, the LLM generates a claim that can be either supported or refuted by the provided document.
By injecting different question types, we prevent the LLM from overfitting to a specific output format and improve the model's generalization ability across different QA tasks.

After generating large amounts of candidate QA pairs, we implement a filtering step to keep only high-quality QA pairs. We define high-quality QA pairs as those that are answerable using the top-$k$ retrieved contexts. Specifically, we retain only those samples where the ground truth answer $a_i'$ is present in the top-$k$ documents retrieved by a strong retriever, such as Dragon~\citep{dragon}, based on the generated query $q_i'$. Formally, the sample is retained if $a_i' \in \mathcal{D}_i'^k$, where $\mathcal{D}_i'^k$ denotes the top-$k$ documents retrieved for query $q_i'$. From these retained samples, we create pseudo-labeled training tuples $\mathcal{T}' = {(q_i', \mathcal{D}_i', a_i')}_{i=1}^n$. 

With the created synthetic tuples $\mathcal{T}'$, we augment it with the SFT data $\mathcal{T}_{\text{SFT}}$ and the general domain context-aware QA data from Stage-I $\mathcal{T}_{\text{gen}}$, to continuously fine-tune our models, enhancing the LLMs' QA abilities within the specific domain. The size and blending ratio of the pseudo-labeled samples can be found in Appendix~\ref{apd:data_ratio}.

\section{Experimental Setup}
\vspace{-0.5ex}
\subsection{Tasks and Datasets}
We evaluate our model across a total of 11 datasets spanning the \textit{medical}, \textit{scientific} and \textit{computer science} domains. 
For the medical domain, we include the five datasets in the MIRAGE benchmark~\citep{xiong-etal-2024-benchmarking}, including PubMedQA~\citep{jin2019pubmedqa}, BioASQ~\citep{bioasq}, MedQA~\citep{medqa}, MedMCQA~\citep{pal2022medmcqa}, the medical subsets in MMLU~\citep{mmlu}, and two additional open-ended QA datasets LiveQA~\citep{abacha2017overview}, and MedicationQA~\citep{abacha2019bridging}.
For the scientific domain, we consider ARC-challenge~\citep{clark2018think}, SciQ~\citep{SciQ}\footnote{We convert the multiple-choice questions in SciQ into short-phrase answer generation tasks to better assess the model's generative capabilities.}, and the scientific subsets (14 subtasks in total) in MMLU~\citep{mmlu}.
For computer science, we use CS-Bench~\citep{song2024cs} for evaluation.
We distinguish the computer science domain from the broader scientific domain as the scientific domain predominantly covers natural and social sciences, with limited representation of computer science topics.
We use accuracy as the evaluation metric for multiple-choice and True-or-False questions, Rouge-L and MAUVE for open-ended questions, Exact Match (EM) and F1 for Fill-in-the-blank questions, with Rouge-L and F1 as the main metrics, respectively. An exception is CS-Bench, where we follow the original paper's evaluation method by using GPT-4 as a judge for fill-in-the-blank and open-ended questions.

For the medical domain, we use the corpora from Textbooks~\citep{medqa}, Wikipedia and PubMed articles\footnote{\url{https://pubmed.ncbi.nlm.nih.gov/}} to generate pseudo-labeled samples in Stage-II. For the scientific domain, we leverage Wikipedia. For the CS domain, we use Wikipedia CS Subset\footnote{\url{https://huggingface.co/datasets/AlaaElhilo/Wikipedia_ComputerScience}} and arXiv articles\footnote{\url{https://huggingface.co/datasets/CCRss/arxiv_papers_cs}}.

\begin{table*}[t]
\centering
\renewcommand\arraystretch{0.93}
\caption{Results of our proposed method and baselines in the medical domain. All the presented methods use RAG for inference. \textbf{Bold} and \underline{underline} highlight the best and second best performance, respectively. $^*$: the main metric used for average calculation. $^\dagger$: models trained using synthetic data generated from GPT-4. $^\ddagger$: our own implementation of the models with the same unlabeled corpora. The notations are the same for the following tables.\vspace{-0.5ex}}
\label{tab:main_med}
\resizebox{\linewidth}{!}{
\begin{tabular}{l|cccccccc}
\toprule
\textbf{Datasets} & \textbf{PubMedQA} & \textbf{BioASQ} & \textbf{MedQA} & \textbf{MedMCQA}  & \textbf{MMLU-med} & \textbf{LiveQA} & \textbf{MedicationQA} 
& \bf Avg. \\
\midrule
\bf Metrics & ACC & ACC & ACC & ACC & ACC & Rouge-L$^*$ / MAUVE & Rouge-L$^*$ / MAUVE & ---\\
\midrule
\multicolumn{3}{l}{\textit{Proprietary LLMs, For Reference Only}} \\
\midrule
GPT-3.5~\citep{chatgpt} & 67.40 & 90.29 & 66.61 & 58.04 & 75.48 & 42.3 / 62.5 & 36.3 / 46.0 & 62.35 \\
GPT-4~\citep{gpt4} & 70.60 & 92.56 & 82.80 & 66.65 & 87.24 & 44.0 / 65.9  & 41.5 / 59.2 & \bf 69.34 \\
\midrule
\multicolumn{3}{l}{\textit{Medical LLMs}} \\
\midrule
PMC-Llama 13B~\citep{pmcllama} & 56.00 & 65.21 & 42.58 & 48.29 & 52.53  & 35.7 / 60.6  & 36.4 / 38.3 & 48.10 \\
MEDITRON 70B~\citep{chen2023meditron} & 56.40 & 76.86 & 49.57 & 52.67 & 65.38 & --- & --- & ---\\
AdaptLLM-v2 8B~\citep{adaptllm} & 45.00 & 78.80 & 43.13 & 42.74 & 51.24 & 30.2 / 48.0 &  39.2 / 51.4 & 47.19 \\
BioMistral 7B~\citep{labrak2024biomistral} & 59.20 & 82.69 & 32.52 & 32.20 & 47.47 & \underline{43.1} / 63.2 & 39.6 / 51.9 & 48.11 \\
MedLlama3 8B~\citep{JSLMedLlama3_8Bv2_0} & 74.20 & 83.50 & 61.43 & 61.18 & 77.13 & 27.9 / 45.2 & 29.8 / 35.0 & 59.31 \\
\midrule
\multicolumn{3}{l}{\textit{Retrieval-Augmented LLMs}} \\
\midrule
Self-RAG 13B$^\dagger$~\citep{asai2024selfrag} & 71.20 & 73.70 & 48.60 & 44.00 & 53.90 & 35.6 / 54.1 & 39.3 / 46.4 & 52.33 \\
ChatQA1.5 8B~\citep{liu2024chatqa} & 66.40 & 82.69 & 42.36 & 46.97 & 61.40 & 39.3 / 65.5 & 39.9 / 48.9 & 54.15 \\
ChatQA1.5 70B~\citep{liu2024chatqa} & 74.80 & 83.17 & 68.89 & 62.54 & 80.51 & 40.1 / 66.3 & \underline{40.8} / 50.2 & 64.40\\
\midrule
\multicolumn{3}{l}{$^\ddagger$\textit{Backbone: Llama3-8B-Instruct}} \\
\midrule
Llama3-8B-it~\citep{llama3} & 64.60 & 88.51 & 55.30 & 58.91 & 69.79 & 34.1 / 54.1 &  37.2 / 45.6 & 58.34 \\
RAFT 8B$^\dagger$~\citep{zhang2024raft} & 73.40 & 88.67 & 54.28 & 60.15 & 70.25 & 36.2 / 55.6 & 38.9 / 56.4 & 60.26 \\
EvidenceRAG 8B$^\dagger$~\citep{synqa} & 75.00 & 90.61 & 57.74 & 61.13 & 72.27 & 36.6 / 57.8 & 34.6 / 53.6 & 61.14 \\
\rowcolor{teal!12} \ours{} 8B & \textbf{80.00} & \underline{91.75} & \underline{62.92} & \textbf{67.51} & 75.57 & \textbf{44.4} / \underline{66.6} & 40.1 / 57.4 & \textbf{66.04} \\
~~~~w/o Stage II & \underline{78.00} & 90.45 & 60.56 & \underline{65.22} & 74.56 & {42.8} / 62.9 & 38.5 / 55.6 & 64.30\\
\midrule
\multicolumn{3}{l}{$^\ddagger$\textit{Backbone: Gemma2-27B-Instruct}} \\
\midrule
Gemma2-27B-it~\citep{team2024gemma} & 56.20 & 89.32 & 59.70 & 57.30 & 75.67 & 37.4 / 52.8 & 40.2 / 57.0 & 59.40\\
RAFT 27B$^\dagger$~\citep{zhang2024raft} & 67.20 & 91.70 & 62.22 & 61.56 & 78.97 & 39.4 / 62.2 & 40.2 / 48.2 & 63.04 \\
EvidenceRAG 27B$^\dagger$~\citep{synqa} & 63.00 & 90.61 & 62.14 & 61.80 & 79.43 & 34.5 / 58.6 & 34.5 / 44.6 & 60.85 \\
\rowcolor{lightcoral!20} \ours{} 27B & 73.60 & \textbf{92.07} & \textbf{63.63} & 64.16 & \textbf{81.63} & 39.9 / \textbf{66.8} & \textbf{41.2} / \textbf{62.1} & \underline{65.17} \\
~~~~w/o Stage II & 66.00 & 91.59 & 62.45 & 58.67 & \underline{79.61} & 37.2 / 61.6 & \underline{40.8} / \underline{58.6} & 62.33 \\

\bottomrule
\end{tabular}
}
\end{table*}

\begin{table}[t]
\centering
\renewcommand\arraystretch{0.95}
\caption{Results of our proposed method and baselines in the scientific domain.\vspace{-0.5ex}}
\label{tab:main_sci}
\resizebox{\linewidth}{!}{
\begin{tabular}{l|cccc}
\toprule
\textbf{Models} & \textbf{MMLU-sci} & \textbf{ARC} & \textbf{SciQ} & \bf Avg. \\
\midrule
\bf Metrics & ACC & ACC & EM / F1$^*$ & ---\\
\midrule
\multicolumn{3}{l}{\textit{Proprietary LLMs, For Reference Only}} \\
\midrule
GPT-3.5~\citep{chatgpt} & 66.40 & 75.30 & 40.30 / 62.73 & 68.14\\
GPT-4~\citep{gpt4} & 87.46 & 94.03 & 43.24 / 66.03 & \bf 82.51 \\
\midrule
\multicolumn{3}{l}{\textit{Scientific LLMs}} \\
\midrule
SciTulu 7B~\citep{wadden2024sciriff} & 55.95 & 53.84 & 22.2 / 40.55 & 50.11  \\
SciTulu 70B~\citep{wadden2024sciriff} & 71.80 & 52.82 & 18.6 / 36.69 & 53.77  \\
\midrule
\multicolumn{3}{l}{\textit{Retrieval-Augmented LLMs}} \\
\midrule
Self-RAG 13B$^\dagger$~\citep{asai2024selfrag} & 48.69 & 73.10 & 31.60 / 51.87 & 57.89  \\
ChatQA 8B~\citep{liu2024chatqa} & 54.46 & 52.22 & 40.40 / 60.60 & 55.76  \\
ChatQA 70B~\citep{liu2024chatqa} & 75.21 & 81.06 & 50.00 / 68.41 & 74.89 \\
\midrule
\multicolumn{3}{l}{$^\ddagger$\textit{Backbone: Llama3-8B-Instruct}} \\
\midrule
Llama3-8B-it~\citep{llama3} & 67.15 & 71.08 & 20.80 / 42.47 & 60.23 \\
RAFT 8B$^\dagger$~\citep{zhang2024raft} & 69.22 & 73.12 & 48.20 / 68.56 & 70.30 \\
EvidenceRAG 8B$^\dagger$~\citep{synqa} & 71.59 & 75.34 & 53.10 / 70.11 & 72.35 \\
\rowcolor{teal!12} \ours{} 8B & 77.31 & 81.40 & \underline{57.50} / \underline{72.17} & 76.96\\
~~~~w/o Stage II & 75.95 & 80.20 & 53.80 / 70.16 & 75.44 \\
\midrule
\multicolumn{3}{l}{$^\ddagger$\textit{Backbone: Gemma2-27B-Instruct}} \\
\midrule
Gemma2-27B-it~\citep{team2024gemma} & 76.11 & 85.75 & 44.80 / 66.99 & 76.28\\
RAFT 27B$^\dagger$~\citep{zhang2024raft} & 78.79 & \underline{86.95} & 53.10 / 70.91 & 78.88 \\
EvidenceRAG 27B$^\dagger$~\citep{synqa} & \underline{78.84} & 86.69 & 45.60 / 67.50 & 77.68\\
\rowcolor{lightcoral!20} \ours{} 27B & \textbf{81.28} & \textbf{88.65} & \textbf{58.10} / \textbf{74.99} & \textbf{81.64} \\
~~~~w/o Stage II & 78.38 & 86.86 & 54.50 / 72.00 & \underline{79.08} \\
\bottomrule
\end{tabular}
}
\end{table}

\begin{table}[t]
\centering
\renewcommand\arraystretch{0.95}
\caption{Results of our proposed method and baselines in the computer science domain. MC, AS, FB, OG stands for multiple-choice, assertion, fill-in-the-blank and Open-ended generation, respectively.}
\label{tab:main_cs}
\resizebox{\linewidth}{!}{
\begin{tabular}{l|ccccc}
\toprule
\textbf{Models} & \textbf{MC} & \textbf{AS} & \textbf{FB} & \textbf{OE} & \bf Overall \\
\midrule
\bf Metrics & ACC & ACC & Auto & Auto & --- \\
\midrule
\multicolumn{5}{l}{\textit{Proprietary LLMs, For Reference Only}} \\
\midrule
GPT-3.5~\citep{chatgpt} & 54.89 & 67.30 & 42.93  & 50.11 & 55.74 \\
GPT-4~\citep{gpt4} & 71.48 & 73.62 & 56.87 & 71.43 & \bf 70.34 \\
\midrule
\multicolumn{5}{l}{\textit{Scientific LLMs}} \\
\midrule
SciTulu 7B~\citep{wadden2024sciriff} & 38.40 & 56.56 & 27.66 & 32.29 & 40.44 \\
SciTulu 70B~\citep{wadden2024sciriff} & 44.24 & 60.18 & 31.06 & 54.76 & 46.87\\
\midrule
\multicolumn{5}{l}{\textit{Retrieval-Augmented LLMs}} \\
\midrule
Self-RAG 13B$^\dagger$~\citep{asai2024selfrag} & 29.87 & 54.52 & 30.64 & 24.94 & 34.56\\
ChatQA 8B~\citep{liu2024chatqa} & 35.33 & 60.18 & 27.66 & 29.82 & 39.11\\
ChatQA 70B~\citep{liu2024chatqa} & 54.94 & 62.67 & 34.89 & 38.53 & 53.07\\
\midrule
\multicolumn{5}{l}{$^\ddagger$\textit{Backbone: Llama3-8B-Instruct}} \\
\midrule
Llama3-8B-it~\citep{llama3} & 52.69 & 60.41 & 26.81 & 44.12 & 50.80 \\
RAFT 8B$^\dagger$~\citep{zhang2024raft} & 54.57 & 60.86 & 32.76 & 40.23 & 52.38 \\
EvidenceRAG 8B$^\dagger$~\citep{synqa} & 54.42 & 62.67 & 35.02  & 42.30 & 53.06\\
\rowcolor{teal!12}\ours{} 8B & 60.63 & 64.93 & 34.47 & 47.11 & 57.63\\
~~~~w/o Stage II & 59.88 & 61.99 & 34.47 & 46.82 & 56.55\\
\midrule
\multicolumn{5}{l}{$^\ddagger$\textit{Backbone: Gemma2-27B-Instruct}} \\
\midrule
Gemma2-27B-it~\citep{team2024gemma} & 59.96 & 62.22 & 40.00 & \textbf{57.50} &58.08 \\
RAFT 27B$^\dagger$~\citep{zhang2024raft} & 60.93 & \underline{66.06} & 39.15 & 53.80 & 59.07\\
EvidenceRAG 27B$^\dagger$~\citep{synqa} & 60.63 & 62.22 & 40.85  & 54.40 & 58.34\\
\rowcolor{lightcoral!20}\ours{} 27B & \textbf{62.87} & \textbf{66.74} & \textbf{43.83}  &  \underline{54.60} &\textbf{60.96} \\
~~~~w/o Stage II & \underline{61.00} & 65.84 & \underline{41.70} & 54.00 & \underline{59.36}\\
\bottomrule
\end{tabular}
}
\end{table}

\vspace{-0.5ex}
\subsection{Baselines}
We categorize our baselines into four groups: (1) \textit{Off-the-shelf general domain LLMs}, which include GPT-3.5~\citep{chatgpt}, GPT-4~\citep{gpt4}, Llama3-8B-it~\citep{llama3}, and Gemma2-27B-it~\citep{team2024gemma}.
(2) \textit{Off-the-shelf domain-specific LLMs}, including PMC-llama-13B~\citep{pmcllama}, MEDITRON-70B~\citep{chen2023meditron}, AdaptLLM-v2-8B~\citep{adaptllm}, BioMistral-7B~\citep{labrak2024biomistral} and MedLlama3-8B~\citep{JSLMedLlama3_8Bv2_0} in the medical domain, as well as 
SciTulu 7B and 70B~\citep{wadden2024sciriff} in both the scientific domain and the computer science domain, due to the absence of LLMs specifically fine-tuned for the computer science domain.
(3) \textit{General domain retrieval-augmented LLMs}, which include Self-RAG-13B~\citep{asai2024selfrag}, ChatQA1.5-8B and 70B~\citep{liu2024chatqa}.
(4) \textit{Domain-specific Retrieval-augmented LLMs}, including RAFT~\citep{zhang2024raft} and EvidenceRAG\footnote{We named this method ourselves, as the model does not have an officially designated name.}~\citep{synqa}. Since RAFT and EvidenceRAG have not released their checkpoints, we re-implemented their methods using the same backbones as our approach. Note that for all the baseline models, we conduct the zero-shot evaluation and augment the context with retrieval for fair comparison.
We also note that we do not compare with several domain-specific baselines such as \citep{zhang2024ultramedical, nori2023can} which have access to task-specific examples that overlap with our evaluation tasks.
\vspace{-1ex}
\subsection{Implementation Details}
We use Llama3-it 8B~\citep{llama3} and Gemma2-it 27B~\citep{team2024gemma} as our backbones. 
For the Gemma-2 model, we use LoRA~\citep{hu2022lora} ($r=32,\alpha=32$) during fine-tuning due to resource constraints.
For both stages, we set the global batch size to 64, with gradient accumulation as 8 and train the model for 1 epoch. For Stage-I, the learning rate is set to $5e-7$ and for Stage-II, it is set to $2e-7$ for the Llama3 backbone and $5e-7$ for the Gemma backbone. 
AdamW optimizer~\citep{loshchilov2018decoupled} is used with $\beta_1=0.9$ and $\beta_2=0.95$. 
To create context-enhanced examples for our synthetic queries, we use Dragon~\citep{dragon} to extend context length for \ours{} and baselines, which improves RAG model robustness~\citep{yu2024rankrag,chain_of_note}. 
For retrieval during evaluation on medical datasets, we follow the original MIRAGE benchmark by using the top-10 retrieval results as context, ensembled from multiple models. For other datasets, we fetch the top-10 passages by Google Search\footnote{\url{https://www.searchapi.io}}. 
All experiments are conducted on 8 NVIDIA A100 GPUs.
The prompt format for answer and question generation and inference can be found in Appendix~\ref{apd:prompt}.
\section{Experimental Results}

\begin{table*}[h!]
\centering
\caption{Performance of SimRAG using Llama-3-8b-it as the backbone and its variants across medical datasets.}
\resizebox{0.98\linewidth}{!}{
\begin{tabular}{@{}lcccccccc@{}}
\toprule
\textbf{Method} & \textbf{PubMedQA} & \textbf{BioASQ} & \textbf{MedQA} & \textbf{MedMCQA} & \textbf{MMLU-med} & \textbf{LiveQA} & \textbf{MedicationQA} & \textbf{Avg.} \\ 
\midrule
\rowcolor{teal!12} SimRAG  8B                            & 80.00          & 91.75          & 62.92          & 67.51          & 75.57          & 44.4          & 40.1          & 66.04          \\  \midrule
SimRAG w/o general SFT data         & 79.60          & 90.78          & 59.47          & 61.92          & 73.09          & 39.9          & 38.9          & 63.38          \\ 
SimRAG w/o general-domain QA        & 80.20          & 91.10          & 61.04          & 65.31          & 72.91          & 42.8          & 39.4          & 64.68          \\ 
SimRAG w/o general retrieval data   & 79.40          & 90.94          & 57.97          & 62.42          & 71.72          & 39.4          & 38.9          & 62.96          \\ 
SimRAG w/o Stage-I                  & 76.80          & 89.81          & 57.97          & 60.02          & 70.71          & 39.3          & 38.5          & 61.87          \\ 
\bottomrule
\end{tabular}
}
\label{tab:simrag-results}
\end{table*}

\subsection{Main Results}

Table~\ref{tab:main_med}, Table~\ref{tab:main_sci}, and Table~\ref{tab:main_cs} present the experimental results for the medical, scientific, and computer science domains, respectively. The results of the 14 tasks in MMLU-sci can be found in Appendix~\ref{apd:add_exp}
From the results, we have the following findings:\\
(1) \ours consistently outperforms baselines across these domains and a variety of question-answering formats. In medical, scientific, and computer science domain, the average performance gain is 8.01\%, 6.37\%, 8.61\% over the Llama variant and 1.19\%, 3.50\%, 3.20\% over the Gemma variant, respectively. 
Besides, \ours{} also achieves comparable performance to strong proprietary models: when using Gemma2-27B as the backbone, we achieve 93.99\%, 98.95\% and 86.66\% of the performance of GPT-4.
This demonstrates the effectiveness and robustness of \ours in adapting general-domain LLMs to specialized domain knowledge using only unlabeled corpora.
\begin{figure*}[t]
    \centering    
    \begin{minipage}{0.74\textwidth}
    \centering
    \includegraphics[width=0.99\textwidth]{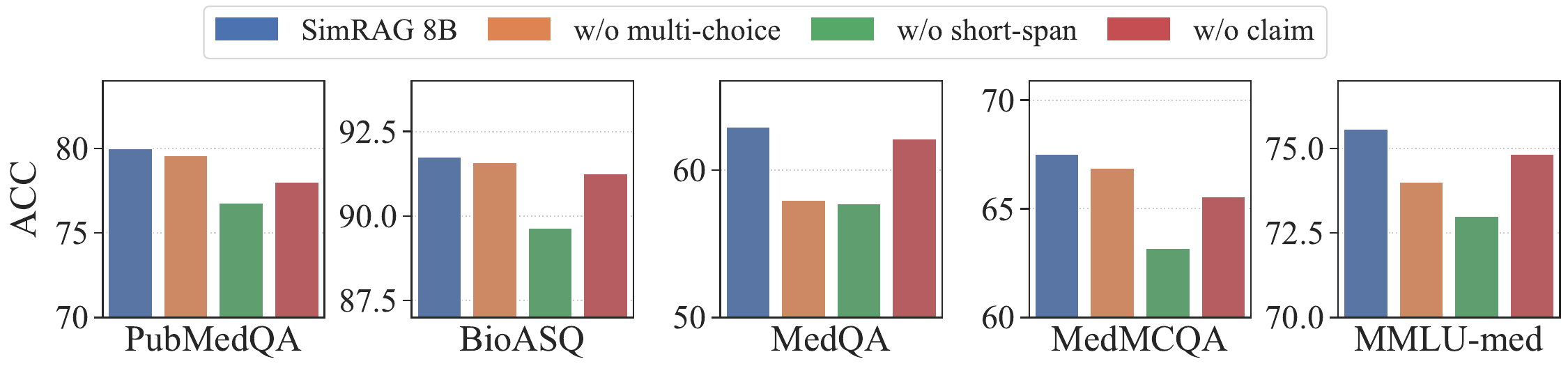}
    \caption{Effect of diverse types of generated QA pairs.}
    \label{fig:diff_components}
    \end{minipage}
    \hfill
    \begin{minipage}{0.24\textwidth}
    \centering
    \includegraphics[width=0.99\textwidth]{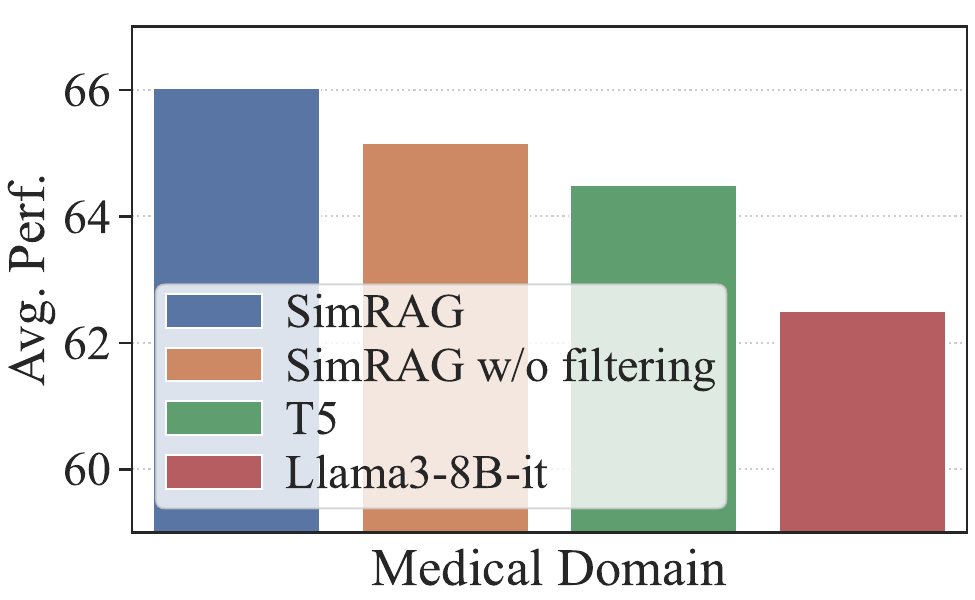}
    \caption{Effect of different generation model.}
    \label{fig:diff_qg}
    \end{minipage}
    \vspace{-1ex}
\end{figure*}

\noindent (2) \emph{Domain-specific LLMs} (e.g. SciTulu and MedLlama), although fine-tuned on relevant data, underperform compared to \ours because they are not optimized for RAG tasks, where effectively utilizing retrieved context is crucial. As a result, they struggle to incorporate relevant context into their answers, leading to weaker performance. On the other hand, \emph{general-domain RAG models} (e.g. ChatQA) face distribution shifts when applied to specialized tasks, as they struggle to  integrate the retrieved domain-specific knowledge accurately. \\
(3) \textit{Domain-specific retrieval-augmented LLMs} such as RAFT and EvidenceRAG still show suboptimal performance despite utilizing the powerful (yet expensive) GPT-4 model to generate synthetic training data.
In contrast, \ours, fine-tuned specifically for the QA generation task, produces more accurate and contextually relevant synthetic QA pairs, leading to better downstream performance across all QA tasks. \\
(4) Although the CS domain is relatively new and less-studied compared to other natural and social sciences, \ours still demonstrates promising performance in this area. This justifies the potential for adapting \ours to emerging domains.

\subsection{Ablation Studies}
\noindent \textbf{Effect of Stage-I and Stage-II.} Table~\ref{tab:main_med} to \ref{tab:simrag-results} show that retrieval-oriented fine-tuning (Stage-I) significantly enhances LLM performance on QA tasks compared to the original backbone, demonstrating its effectiveness. 
However, further improvements become challenging after this stage. When the LLMs are fine-tuned on self-synthesized training tuples, their performance on target tasks improves even more, with an average increase of 2.21\% for Llama and 3.50\% for Gemma.. This suggests that, with access to a target domain corpus, LLMs can generate high-quality synthetic data, enabling self-improvement and further boosting performance.

\begin{table}[t]
\centering
\caption{Results of the 5 datasets from the medical MIRAGE benchmark~\citep{xiong-etal-2024-benchmarking}, using DRAGON~\citep{dragon} as an alternative retriever.}
\label{tab:diff_retriever}
\resizebox{\linewidth}{!}{
\begin{tabular}{l|cccccc}
\toprule
\textbf{Models} & \textbf{PubMedQA} & \textbf{BioASQ} & \textbf{MedQA} & \textbf{MedMCQA}  & \textbf{MMLU-med} & \bf Avg. \\
\midrule
Llama3-8B-it~\shortcite{llama3} & 57.00 & 81.55 & 55.70 & 55.16 & 65.93 & 63.07\\
\rowcolor{teal!12}\ours 8B & \textbf{79.60} & \textbf{91.42} & \underline{60.80} & \textbf{63.88} & 74.01 & \textbf{73.94}\\
\midrule
Gemma2-27B-it~\shortcite{team2024gemma} & 58.80 & 89.48 & 57.97 & 55.13 & \underline{76.67} & 67.61\\
\rowcolor{lightcoral!20}\ours 27B & \underline{73.60} & \underline{90.94} & \textbf{62.29} & \underline{60.39} & \textbf{79.06} & \underline{73.26}\\
\bottomrule
\end{tabular}
}
\end{table}

\noindent \textbf{Effect of Different Retrievers.}  We show the performance of \ours{} using Dragon~\citep{dragon} as the retriever in Table~\ref{tab:diff_retriever}. 
The results show consistent performance improvements of \ours{} over the LLM backbone, demonstrating that \ours{} is robust to different retriever choices and that its self-improvement mechanism consistently enhances performance.

\begin{table*}[t]
\centering
\caption{Two case studies comparing the generated pseudo-labeled QA pairs from our fine-tuned model after Stage-I with those generated by the backbone model, Llama3-8B-it. \textcolor{green}{Green} and \textcolor{red}{red} texts represent the corresponding information between the generated QA pairs and the ground truth context.}
\label{tab:case_study}
\resizebox{\linewidth}{!}{
    \begin{tabular}{p{2.5cm}p{22.5cm}}
       \toprule
        \multicolumn{2}{c}{\bf Case Study 1} \\
       \midrule
       \bf Corpus / Type & Textbooks / Claim Verification\\
       \midrule 
        \bf Context & \textbf{Title:} Histology Ross, \textbf{Text:} For its survival, the body needs to ensure a continuous delivery of energy despite highly variable supplies of nutrients from the external environment. To meet the body's energy demands when nutrient supplies are low, \textcolor{green}{adipose tissue} efficiently stores excess energy. \textcolor{green}{The body} has a limited capacity to store carbohydrate and protein, therefore \textcolor{green}{energy} reserves are stored within lipid droplets of adipocytes in the form of \textcolor{green}{triglycerides}. Triglycerides represent a dynamic form of energy storage that is added to when food intake is greater than energy expenditure and is tapped when energy expenditure is greater than food intake. \textcolor{red}{The energy stored in adipocytes can be rapidly released for use at other sites in the body.} \\
        \midrule
        \bf Llama3-8B-it & \textbf{Claim:} \textcolor{red}{The energy stored in adipocytes can be rapidly released for use at other sites in the body.} \newline \textbf{Label:} Supported. \\
        \midrule
        \bf \ours \newline (after Stage-I) & \textbf{Claim:} \textcolor{green}{The body} stores \textcolor{green}{energy} in the form of \textcolor{green}{triglycerides} in \textcolor{green}{adipose tissue}. \newline \textbf{Label:} Supported.\\
       \toprule
       \multicolumn{2}{c}{\bf Case Study 2} \\
       \midrule
       \bf Corpus / Type & Wikipedia the medical subset / Short-span QA\\
       \midrule 
        \bf Context & \textbf{Title:} Cognitive behavioral therapy, \textbf{Text:} Emerging evidence for cognitive behavioral interventions aimed at \textcolor{green}{reducing symptoms} of depression, \textcolor{green}{anxiety}, and obsessive-compulsive disorder in autistic adults without intellectual disability has been identified through a systematic review. While the research was \textcolor{red}{focused on adults}, cognitive behavioral interventions have also been beneficial to autistic children. \\
        \midrule
        \bf Llama3-8B-it & \textbf{Question:} What age group of individuals with autism spectrum disorder, who do not have intellectual disability, have been found to benefit from cognitive behavioral interventions aimed at reducing symptoms of depression, anxiety, and obsessive-compulsive disorder, and are also a \textcolor{red}{focus of}? \newline \textbf{Answer:} \textcolor{red}{children}.\\
        \midrule
        \bf \ours \newline (after Stage-I) & \textbf{Question:} What is one of the disorders that cognitive behavioral therapy can \textcolor{green}{help with}? \newline \textbf{Answer:} \textcolor{green}{anxiery}.\\
       \bottomrule
    \end{tabular}
}
\end{table*}

\subsection{Study on Pseudo-labeled Tuples}
We mainly demonstrate the advantage of \ours{} in generating pseudo-labeled data from the following three perspectives.

\noindent \textbf{Effect of different question generation models.}
To demonstrate the benefit of training on question generation and question-answering data, we compare the performance of Stage-II using different synthetic question-answer pairs. These pairs are generated either directly by Llama-3-8b-it or by an off-the-shelf QG model with T5-Large~\cite{raffel2020exploring} as the backbone. The results demonstrate that our approach achieves better performance on average, demonstrating the clear advantage of leveraging the fine-tuned model itself for pseudo-labeled data generation.

\noindent \textbf{Effect of question filtering.}
We further demonstrate the advantages of question filtering in Figure~\ref{fig:diff_qg}, showing that removing low-quality data not only improves overall model performance but also accelerates the training process.
It is also worth noting that even without filtering, \ours{} can achieve strong performance, suggesting that the synthetic questions generated from fine-tuned LLMs are already highly relevant to the context.

\noindent \textbf{Effect of diverse question types.}
From Figure~\ref{fig:diff_components}, we observe that \ours{} achieves the best performance when all three different types are included. These results justify the necessity for incorporating different task types into the fine-tuning step in Stage-II. Besides, claim verification benefits PubMedQA and BioASQ more, while multiple-choice questions boost performance on MedQA, MedMCQA, and MMLU, aligning with the question types in each dataset. Lastly, we observe that removing short-span QA leads to the largest performance drops, indicating its central role in adapting the LLM's performance towards specialized domains.

\subsection{Case Studies}
To better illustrate the quality of pseudo-labeled samples generated by \ours after Stage-I fine-tuning, we present two case studies in Table~\ref{tab:case_study}, comparing the samples produced by \ours with those from the baseline model, Llama3-8B-it.

In the first case, where the model is asked to generate a claim supported by the context, Llama3-8B-it simply selects a sentence  from the context. This results in relatively simple QA pairs, making the task less challenging for Stage-II training.

In the second case, the model is tasked with generating an answer first, and then formulating a question based on the context and the answer. While Llama3-8B-it does not copy a sentence exactly, it generates a lengthy question that closely paraphrases the context. This makes the question overly dependent on the original text, making it difficult to interpret without it. Additionally, the model misinterprets the context by implying that the research was focused on children when actually adults are the focus.
In contrast, after fine-tuning on answer generation and query generation in Stage-I, \ours generates higher-quality QA pairs that are self-contained and understandable without relying on the context. These QA pairs also present more challenging tasks, as they require deeper comprehension of the context, providing harder and more effective training data for Stage-II.

\section{Conclusion}
We introduce \ours{}, an instruction fine-tuning framework designed to enhance LLMs for domain-specific question-answering tasks. By equipping LLMs with joint capabilities for both question answering and question generation, \ours{} enables the generation of diverse, high-quality synthetic questions from unlabeled domain-relevant corpora. This approach facilitates effective adaptation to specialized fields, where distribution shifts and limited domain-specific data typically pose challenges.
Extensive experiments across 11 datasets in three domains show that \ours{} consistently outperforms baseline models, demonstrating its effectiveness in tackling the challenges of retrieval-augmented, domain-specific question-answering tasks.

\section*{Limitation}
While \ours{} demonstrates notable improvements, there are some limitations to our approach:

\noindent \textbf{Single Round Pseudo-Label Generation}: Our current method relies on a single round of query generation from the corpus, which may restrict the refinement of pseudo label quality. Iterative refinement of generated synthetic queries could potentially lead to better results.

\noindent \textbf{Additional Training Time}: The incorporation of synthetic query generation and filtering adds time complexity compared to baseline models, which may affect efficiency in environments with limited computational resources. However, we would like to note that our method \emph{will not increase the inference time complexity} compared to the existing RAG approaches with the same backbone models.

\noindent \textbf{Stronger Query Generation Models}: Although we achieved strong performance with Llama3 8B and Gemma2 27B models, leveraging more powerful query generation models, such as Llama-3.1-70B-it~\citep{llama3}, could yield further gains. However, using larger models would incur higher computational costs beyond our current budget.

\bibliography{anthology,custom}

\newpage
\appendix

\begin{table*}[!t]
\centering
\caption{The blending ratio of different datasets with their specific prompt format in Stage-I and Stage-II fine-tuning. For Stage-II Pseudo-labeled QA Samples, the two numbers represent the \# sample for the Llama and Gemma backbones, respectively.}
\label{tab:blending_ratio}
\resizebox{\linewidth}{!}{
\begin{tabular}{l|>{\centering}m{10cm}|cccc}
\toprule
\multirow{2}{*}{\textbf{Dataset}} & \multirow{2}{*}{\textbf{Specific Instruction}} & \textbf{Stage-I} & \textbf{Stage-I} & \textbf{Stage-II}  & \textbf{Stage-II} \\
& & \textbf{\# Samples} & \textbf{Blending Ratio} & \textbf{\# Samples}  & \textbf{Blending Ratio} \\
\midrule
\rowcolor{gray!12} \multicolumn{6}{l}{\textbf{Instruction Fine-tuning}} \\
\midrule
ChatQA SFT Data & --- & 60000 & 0.18 & 128000 & 0.12\\
\midrule
\rowcolor{gray!12} \multicolumn{6}{l}{\textbf{Question Answering}} \\
\midrule
DROP & \multirow{8.5}{*}{Answer the following question with a short span.} & 12000 & 0.034 & 29195 & 0.04\\
\cmidrule(lr){1-1} \cmidrule(lr){3-6}
NarrativeQA & & 12000 & 0.034 & 40000 & 0.04\\
\cmidrule(lr){1-1} \cmidrule(lr){3-6}
Quoref & & 4800 & 0.014 & 10996 & 0.015\\
\cmidrule(lr){1-1} \cmidrule(lr){3-6}
ROPES & & 4800 & 0.014 & 10924 & 0.015\\
\cmidrule(lr){1-1} \cmidrule(lr){3-6}
Squad1.1 & & 16000 & 0.045 & 40000 & 0.035\\
\cmidrule(lr){1-1} \cmidrule(lr){3-6}
Squad2.0 & & 16000 & 0.045 & 52474 & 0.05\\
\midrule
\centering OpenbookQA & \multirow{2.5}{\linewidth}{\centering Answer the following question by selecting one of the provided options with A, B, C, or D. Please answer with the capitalized alphabet only, without adding any extra phrase or period.} & 2000 & 0.006 & 82092 & 0.005\\ [0.8ex]
\cmidrule(lr){1-1} \cmidrule(lr){3-6}
LogiQA & & 4000 & 0.012 & 7376 & 0.006\\ [0.8ex]
\midrule
NQ & Answer the following question with a short phrase. & 16000 & 0.045 & 46426 & 0.04\\
\midrule
TatQA-arithmetic & Answer the following question with a number from context or the math arithmetic using +,-,*, or /. & 8325 & 0.045 & 24975 & 0.034\\
\midrule
TatQA-others & Answer the following question with a short span, or a full and complete answer. & 3176 & 0.023 & 9528 & 0.013\\
\midrule
WebGLM & Please give a full and complete answer for the question using only the provided search results (some of which might be irrelevant) and cite them properly. Use an unbiased and journalistic tone. When citing several search results, use [1][2][3]. & 12000 & 0.034 & 43579 & 0.023\\
\midrule
StrategyQA & \multirow{2}{*}{Answer the following question with Yes or No.} & 1526 & 0.005 & 4578 & 0.006\\
\cmidrule(lr){1-1} \cmidrule(lr){3-6}
BoolQ & & 4000 & 0.012 & 9427 & 0.013\\
\midrule
FaVIQ & \multirow{2}{\linewidth}{\centering Answer the following question with Yes or No. Is the statement \{claim\} correct?} & 2000 & 0.006 & 10906 & 0.01\\
\cmidrule(lr){1-1} \cmidrule(lr){3-6}
FEVER & & 2000 & 0.006 & 10444 & 0.01\\
\midrule
\rowcolor{gray!12} \multicolumn{6}{l}{\textbf{Pseudo-labeled Question Answering}} \\
\midrule
Short-span QA  & Answer the following question with a short span. & --- & --- & 150,000 / 45,000 & 0.2625 \\
\midrule
Multiple-choice QA & Answer the following question by selecting one of the provided options with A, B, C, or D. Please answer with the capitalized alphabet only, without adding any extra phrase or period. & --- & --- & 50,000 / 15,000 & 0.0875\\
\midrule
Claim Verification & Answer the following question with Yes or No. Is the statement \{claim\} correct? & --- & --- & 100,000 / 30,000 & 0.175 \\
\midrule
\rowcolor{gray!12} \multicolumn{6}{l}{\textbf{Answer Generaion}} \\
\midrule
Squad1.1 & \multirow{5}{\linewidth}{\centering Based on the context, generate candidate spans within the passage that are likely to be answers to a question. Separate different candidate answers with a semicolon (';').} & 18877 & 0.063 & --- & --- \\
\cmidrule(lr){1-1} \cmidrule(lr){3-6}
Squad2.0 & & 18863 & 0.059 & --- & --- \\
\cmidrule(lr){1-1} \cmidrule(lr){3-6}
DROP & & 4984 & 0.023 & --- & --- \\
\cmidrule(lr){1-1} \cmidrule(lr){3-6}
WebQuestions & & 1084 & 0.012 & --- & --- \\
\midrule
\rowcolor{gray!12} \multicolumn{6}{l}{\textbf{Query Generaion}} \\
\midrule
NQ & \multirow{5}{\linewidth}{\centering Based on the context, please generate a question. The answer to the question should be \{answer\}.} & 20000 & 0.068 & --- & --- \\
\cmidrule(lr){1-1} \cmidrule(lr){3-6}
Squad1.1 & & 20000 & 0.068 & --- & --- \\
\cmidrule(lr){1-1} \cmidrule(lr){3-6}
StrategyQA & & 131 & 0.023 & --- & --- \\
\cmidrule(lr){1-1} \cmidrule(lr){3-6}
WebQuestions & & 24000 & 0.068 & --- & --- \\
\midrule
FaVIQ & \multirow{2.5}{\linewidth}{\centering Based on the context, please generate a claim that can be {supported/refuted} by the context.} & 10000 & 0.028 & --- & --- \\
\cmidrule(lr){1-1} \cmidrule(lr){3-6}
FEVER & & 10000 & 0.028 & --- & --- \\
\bottomrule
\end{tabular}
}
\end{table*}

\section{Training Data Details}
\label{apd:data_ratio}
We include the training dataset, the number of examples used in each stage, as well as the instruction format in Table~\ref{tab:blending_ratio}.

\section{Test Data Details}
We evaluate on 11 datasets in total from the medical, scientific and computer science domain.
(1) \textbf{Medical:}
\begin{itemize}[leftmargin=*]
    \item MMLU-med~\citep{mmlu} is a subset of six tasks related to biomedicine, including anatomy, clinical knowledge, professional medicine, human genetics, college medicine, and college biology. It contains 1089 questions in total.
    \item MedMCQA~\citep{pal2022medmcqa} includes multiple-choice questions derived from Indian medical entrance exams, covering 2400 healthcare topics across 21 medical subjects. We use the 4,183-question development set from MedMCQA, as the test set lacks provided ground truths.
    \item MedQA~\citep{medqa} is collected from the US Medical Licensing Examination, containing 1273 four-option multiple-choice questions focused on real-world scenarios from professional medical board exams.
    \item BioASQ~\citep{bioasq} includes 618 questions constructed from biomedical literature without providing the ground truth snippets, challenging RAG systems to infer answers independently.
    \item PubMedQA~\citep{jin2019pubmedqa} is a biomedical research QA dataset consisting of 1000 manually annotated questions based on PubMed abstracts. Answers in PubMedQA are structured as yes/no/maybe to reflect the validity of the questions.
    \item LiveQA~\citep{abacha2017overview} and MedicationQA~\citep{abacha2019bridging} are two QA datasets focusing on answering consumer health questions about medications, including 100 and 674 question-answer pairs, respectively.
\end{itemize}

(2) \textbf{Scientific:}
\begin{itemize}[leftmargin=*]
    \item SciQ~\citep{SciQ} is a scientific question-answering dataset containing 13,679 crowdsourced science exam questions about Physics, Chemistry, and Biology, among others.
    \item ARC-easy/challenge~\citep{clark2018think} contains 7,787 authentic multiple-choice science questions at the grade-school level, designed to foster advanced question-answering research. The dataset is divided into a Challenge Set, with questions that stumped both a retrieval-based and a word co-occurrence algorithm, and an Easy Set.
    \item MMLU-Sci~\citep{mmlu} is the Massive Multitask Language Understanding dataset, designed to test a wide range of language understanding abilities across 57 tasks. In this work, we select 14 subjects to ensure the evaluation is not limited to certain fields.
\end{itemize}

(3) \textbf{Computer Science:}
\begin{itemize}[leftmargin=*]
    \item  CS-Bench~\citep{song2024cs} is a recently-proposed benchmark specifically designed to assess the performance of large language models (LLMs) in computer science.  It contains around 5,000 carefully selected test samples that span 26 subfields within four major areas of computer science, covering various task forms and divisions of knowledge and reasoning. 
\end{itemize}


\section{Baseline Descriptions}
    \begin{itemize}[leftmargin=*]
    
        \item Self-RAG~\citep{asai2024selfrag} utilizes instruction fine-tuning to adaptively retrieve passages based on the question and determine if the passage contains useful information for answering the question. 
        \item ChatQA~\citep{liu2024chatqa} is a fine-tuning pipeline tailored for RAG and conversational QA tasks via aggregating multiple QA and dialogue datasets.
        \item RAFT~\citep{zhang2024raft} is a domain-specific fine-tuning approach that incorporates top-$k$ passages as context during fine-tuning, helping to address discrepancies between training and testing data.
        \item EvidenceRAG~\citep{synqa} leverage off-the-shelf LLMs (GPT-4) to generate context-aware question answering datasets, which is then used to fine-tune the student model.  
        
    \end{itemize}

\begin{table*}[t]
\centering
\caption{Results of our proposed method and baselines in the scientific domain.}
\label{tab:main_sci_mmlu}
\resizebox{\linewidth}{!}{
\begin{tabular}{l|ccccccccccccccc}
\toprule
\bf \multirow{2}{*}{Models} & \bf \multirow{2}{*}{astronomy} & \bf college & \bf college & \bf college & \bf computer & \bf high school & \bf high school & \bf high school & \bf high school & \bf high school & \bf high school & \bf human & \bf \multirow{2}{*}{nutrition} & \bf \multirow{2}{*}{virology} & \bf \multirow{2}{*}{Avg.} \\
& & \bf biology & \bf chemistry & \bf physics & \bf security & \bf geography & \bf macroeconomics & \bf microeconomics & \bf psychology & \bf US history & \bf world history & \bf sexuality & \\
\midrule
\bf Metrics & ACC & ACC & ACC & ACC & ACC & ACC & ACC & ACC & ACC & ACC & ACC & ACC & ACC & ACC & ---\\
\midrule
\multicolumn{3}{l}{\textit{Proprietary LLMs, For Reference Only}} \\
\midrule
GPT-3.5~\citep{chatgpt} & 66.45 & 65.28 & 35.00 & 46.53 & 65.00 & 77.27 & 91.54 & 64.29 & 83.12 & 78.43 & 72.15 & 70.99 & 66.01 & 47.59 & 66.40 \\
GPT-4~\citep{gpt4} & 93.42 & 93.75 & 61.00 & 73.27 & 91.00 & 94.95 & 97.95 & 94.54 & 96.15 & 95.59 & 94.51 & 93.13 & 89.22 & 56.02 & 87.46 \\
\midrule
\multicolumn{3}{l}{\textit{Scientific LLMs}} \\
\midrule
SciTulu 7B~\citep{wadden2024sciriff} & 69.74 & 63.89 & 31.00 & 18.63 & 62.00 & 70.20 & 56.58 & 57.08 & 77.43 & 53.06 & 57.38 & 65.65 & 54.90 & 45.78 & 55.95 \\
SciTulu 70B~\citep{wadden2024sciriff} & 83.55 & 80.56 & 36.00 & 28.43 & 83.00 & 89.39 & 80.26 & 79.83 & 91.19 & 77.55 & 77.22 & 78.63 & 68.95 & 50.60 & 71.80 \\ 
\midrule
\multicolumn{3}{l}{\textit{Retrieval-Augmented LLMs}} \\
\midrule
Self-RAG 13B~\citep{asai2024selfrag} & 55.26 & 58.33 & 24.00 & 21.57 & 60.00 & 61.11 & 32.89 & 45.49 & 67.89 & 58.67 & 58.23 & 53.44 & 43.79 & 40.96 & 48.69 \\
ChatQA 8B~\citep{liu2024chatqa} & 60.53 & 54.17 & 29.00 & 33.33 & 70.00 & 64.65 & 51.32 & 58.37 & 74.86 & 49.49 & 54.85 & 59.54 & 57.19 & 45.18 & 54.46 \\
ChatQA 70B~\citep{liu2024chatqa} & 82.89 & 79.17 & 46.00 & 48.04 & 83.00 & 84.85 & 80.26 & 84.98 & 91.74 & \textbf{86.73} & 82.28 & 74.05 & 77.78 & 51.20 & 75.21 \\
\midrule
\multicolumn{3}{l}{\textit{Backbone: Llama3-8B-Instruct}} \\
\midrule
Llama3-8B-it~\citep{llama3} & 78.29 & 71.53 & 38.00 & 40.20 & 83.00 & 82.32 & 63.16 & 72.96 &  84.04 & 65.31 & 72.15 & 69.47 & 70.26 & 49.40 & 67.15 \\
RAFT 8B~\citep{zhang2024raft} & 80.26 & 75.69 & 37.00 & 42.16 & 84.00 & 79.80 & 65.79 & 74.68 & 83.67 & 72.45 & 77.22 & 73.28 & 71.24 & 51.81 & 69.22 \\
EvidenceRAG 8B~\citep{synqa} & 77.63 & 78.47 & 44.00 & 45.10 & 85.00 & 84.85 & 72.37 & 74.68 & 86.24 & 74.49 & 79.32 & 74.05 & 74.84 & 51.20 & 71.59 \\
\rowcolor{teal!12} \ours{} 8B & \underline{85.53} & 81.94 & \underline{47.00} & 50.98 & \textbf{88.00} & 89.90 & 76.32 & 84.55 & 92.66 & 83.16 & 81.43 & \underline{84.73} & \textbf{81.37} & \underline{54.82} & 77.31\\
~~~~~w/o Stage II & 84.87 & 81.25 & \textbf{49.00} & 49.02 & \underline{87.00} & 88.89 & 73.68 & 82.83 & 90.64 & 80.61 & 81.01 & 83.21 & \underline{79.41} & 51.81 & 75.95 \\
\midrule
\multicolumn{3}{l}{\textit{Backbone: Gemma2-27B-Instruct}} \\
\midrule
Gemma2-27B-it~\citep{team2024gemma} & 82.89 & 84.03 & \underline{47.00} & 55.88 & 84.00 & 89.39 & 77.63 & 81.12 & 91.93 & 80.61 & 84.81 & 81.68 & 72.22 & 52.40 & 76.11 \\
RAFT 27B~\citep{zhang2024raft} & 84.87 & \underline{88.89} & \underline{47.00} & \underline{63.73} & 86.00 & 90.91 & \textbf{86.84} & 84.55 & 93.58 & 81.12 & 85.65 & 81.68 & 76.47 & 51.81 & 78.79 \\
EvidenceRAG 27B~\citep{synqa} & 84.87 & 87.50 & \textbf{49.00} & 60.78 & 86.00 & \underline{91.41} & \textbf{86.84} & \underline{85.41} & \underline{93.94} & 81.63 & 86.08 & 81.68 & 76.80 & 51.81 & \underline{78.84}\\
\rowcolor{lightcoral!20} \ours{} 27B & \textbf{90.13} & \textbf{91.67} & \textbf{49.00} & \textbf{68.63} & \underline{87.00} & \textbf{92.42} & \underline{85.53} & \textbf{87.98} & \textbf{95.05} & \underline{84.18} & \textbf{86.92} & \textbf{85.50} & 78.43 & \textbf{55.42} & \textbf{81.28}\\
~~~~~w/o Stage II & 84.21 & 87.50 & \textbf{49.00} & 59.80 & 84.00 & 89.90 & 84.21 & 82.83 & 93.58 & 83.16 & \underline{86.50} & 81.68 & 76.14 & \underline{54.82} & 78.38 \\
\bottomrule
\end{tabular}
}
\end{table*}

\section{Additional Experimental Results}
\label{apd:add_exp}
We list the per-task results of MMLU-sci in Table~\ref{tab:main_sci_mmlu}.

\section{Prompt Details}
\label{apd:prompt}
\subsection{Answer Generation}
\begin{lstlisting}[style=mystyle, label=lst:prompt, escapeinside={<@}{@>}]
<@\textcolor{blue}{[System]}@>

<@\textcolor{blue}{[Context]}@>

Based on the context, generate several candidate spans within the passage that are likely to be answers to a question. The answers can be entities, verbs or even numbers. Make sure that the answers are different and diverse. Separate different candidate answers with a semicolon (';').
\end{lstlisting}

\subsection{Query Generation}
\begin{lstlisting}[style=mystyle, label=lst:prompt, escapeinside={<@}{@>}]
<@\textcolor{blue}{[System]}@>

<@\textcolor{blue}{[Context]}@>

Based on the context, please generate a question that is relevant to the information provided. The question should stand alone and not refer back to the context explicitly. The question should be clear and understandable without needing the context. The answer to the question should be <@\textcolor{blue}{[Answer]}@>.
\end{lstlisting}

\subsection{Inference}
\begin{lstlisting}[style=mystyle, label=lst:prompt, escapeinside={<@}{@>}]
<@\textcolor{blue}{[System]}@>

<@\textcolor{blue}{[Top 10 Contexts]}@>

<@\textcolor{blue}{[Specific Instruction]}@>

<@\textcolor{blue}{[Question]}@>
\end{lstlisting}

The \textcolor{blue}{[Specifc Instruction]} for each evaluation dataset depends on their question type and can refer to those in Table~\ref{tab:blending_ratio}.

\end{document}